\documentclass[10pt,twocolumn,letterpaper]{article}

\usepackage{iccv}
\usepackage{times}
\usepackage{epsfig}
\usepackage{graphicx}
\usepackage{amsmath}
\usepackage{amssymb}
\usepackage{algorithm}
\usepackage{algpseudocode}

\usepackage[pagebackref=true,breaklinks=true,letterpaper=true,colorlinks,bookmarks=false]{hyperref}

\iccvfinalcopy 

\def\iccvPaperID{15} 

\ificcvfinal\fi
\usepackage{array}
\newcolumntype{M}[1]{>{\centering\arraybackslash}m{#1}}

\begin{document}

\title{Eye Semantic Segmentation with A Lightweight Model}

\author{Van Thong Huynh, Soo-Hyung Kim\thanks{Corresponding author} , Guee-Sang Lee,  Hyung-Jeong Yang\\
School of Electronics and Computer Engineering\\Chonnam National University\\
Gwangju, Korea\\
{\tt\small hvthong.298@outlook.com.vn, \{shkim,gslee,hjyang\}@jnu.ac.kr}
}

\maketitle
\ificcvfinal\thispagestyle{empty}\fi

\begin{abstract}
   In this paper, we present a multi-class eye segmentation method that can run the hardware limitations for real-time inference. Our approach includes three major stages: get a grayscale image from the input, segment three distinct eye region with a deep network, and remove incorrect areas with heuristic filters. Our model based on the encoder-decoder structure with the key is the depthwise convolution operation to reduce the computation cost. We experiment on OpenEDS, a large scale dataset of eye images captured by a head-mounted display with two synchronized eye facing cameras. We achieved the mean intersection over union (mIoU) of $94.85\%$ with a model of size $0.4$ megabytes.
\end{abstract}

\section{Introduction}

Understanding the human eye is an active research because of its significant role in many fields such as psychology, human-computer interaction. Through the years, many researches have been done to segment eye regions in the image.
 
Iris segmentation has been drawing significant attention from the research community due to the popularity of iris recognition technology. In~\cite{sankowski2010reliable}, the authors presented an algorithm that segmented iris in color eye images taken under visible and near-infrared light. It analyzed the color of image, which consisted of four stages: reflection localization, reflections filling in, localize iris boundaries, and eyelid boundary localization. Conversely, ATTention U-Net (ATT-UNet), a method based on feature learning, which guided the model to learn more discriminative features for separating the iris and non-iris pixels proposed in~\cite{lian2018attention}. ATT-UNet deployed a bounding box regression to generate an attention mask for iris, which used as a weighted function to make the model pay more attention to the iris region.
 
Sclera segmentation is typically considered a subproblem of a broader task~\cite{radu2015robust} such as iris recognition or gaze estimation. In~\cite{naqvi2019sclera}, the author presented Sclera-Net, a residual encoder-decoder network based on SegNet, to segment the sclera in various sensor images. The authors in~\cite{lucio2018fully} proposed a method with two steps: periocular region localization, sclera segmentation in the detected region based on Fully Convolutional Network and Generative Adversarial Network. In the case of ~\cite{sclera2019segnet}, they achieved the best performance for sclera segmentation by improving U-Net with channel-wise attention.
 
In multi-class eye segmentation,~\cite{rot2018deep} trained a convolutional encoder-decoder network with 4-fold cross-validation on a small dataset of 120 images from 30 participants. A study based on Atrous convolutional with the conditional random field for post-processing presented in~\cite{luo2019ibug}.

\section{Approach}

In this work, we utilize a Convolutional Encoder-Decoder architecture to segment the 2D grayscale eye image into three distinct classes: the sclera, pupil, and iris. A heuristic filter is performed to reduce the false positive from the network output.
\subsection{Encoder-Decoder architecture}
\begin{figure}
    \centering
    \includegraphics[width=.75\linewidth]{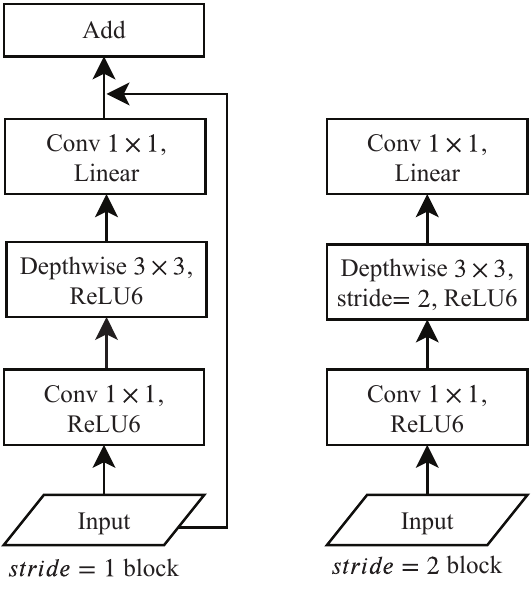}
    \caption{MobileNetV2 bottleneck.}
    \label{fig:mnv2}
\end{figure}

We deployed our encoder based on the bottleneck building block in MobileNetV2~\cite{sandler2018mobilenetv2}. It is a combination of depth-separable convolution, a factorized version of standard convolution, with residuals as in \autoref{fig:mnv2}. Depth separable convolution splits convolution into two separate layers: depthwise convolution for spatial filtering and $1\times1$ pointwise convolutions to generate new features by computing linear combinations of the output from the depthwise layer. This factorization reduces computation by approximate a factor of $k^{2}$~\cite{sandler2018mobilenetv2} compared to the traditional convolution. The architecture of our encoder includes a fully convolution with 32 and 64 filters for the initial and the last layer, 9 residual bottleneck layers inserted in the middle as in \autoref{tab:encoder_info}. We use kernel size $3\times3$ and utilize batch normalization during training.

\begin{table}[htbp]
    \begin{center}
    \begin{tabular}{|c | c |c |c |c |c |} \hline
         Input & Operator & $t$ & $c$ & $n$ & $s$  \\ \hline\hline
         $320\times200$& conv2d & - & 32 & 1 & 2 \\
         $320\times200$ & bottleneck & 1 & 16 & 2 & 1 \\
         $160\times 100$ & bottleneck & 6 & 24 & 3 & 2 \\
         $80\times 50$& bottleneck & 6 & 32 & 4 & 2\\
         $40\times 25$& conv2d & - & 64 & 1 & 1 \\ \hline
    \end{tabular}
	\end{center}
    \caption{The encoder architecture of our method.}\label{tab:encoder_info}
\end{table}

In the decoder module, we build an architecture in which its structure is similar to the Squeeze and Excitation (SE) block in~\cite{hu2018squeeze}. The goal of SE block is to acquire the global information to selectively emphasize informative features and suppress less useful ones by explicitly modeling the interdependencies between channels~\cite{hu2018squeeze}. Our decoder initial with a regular component in the segmentation network: a convolution with 64 filters of kernel size $3\times3$ followed by a bilinear upsampling which increases the input size four times. At this point, we create two different streams to learn and make an ensemble of them at the end of the decoder module. In the first stream, we use three convolutions with kernel size $3\times3$ followed by a bilinear upsampling to get the same size with the original input size. In the other stream, we use only a $1\times1$ convolution and upsampling the output. In order to learn vital information and eliminate the trivial ones, we set the reduction ratio of $r$ to $4$ in the SE block. After each convolution operation, we always use batch normalization and ReLU as the non-linearity function.
\begin{figure}[htbp]
    \centering
    \includegraphics{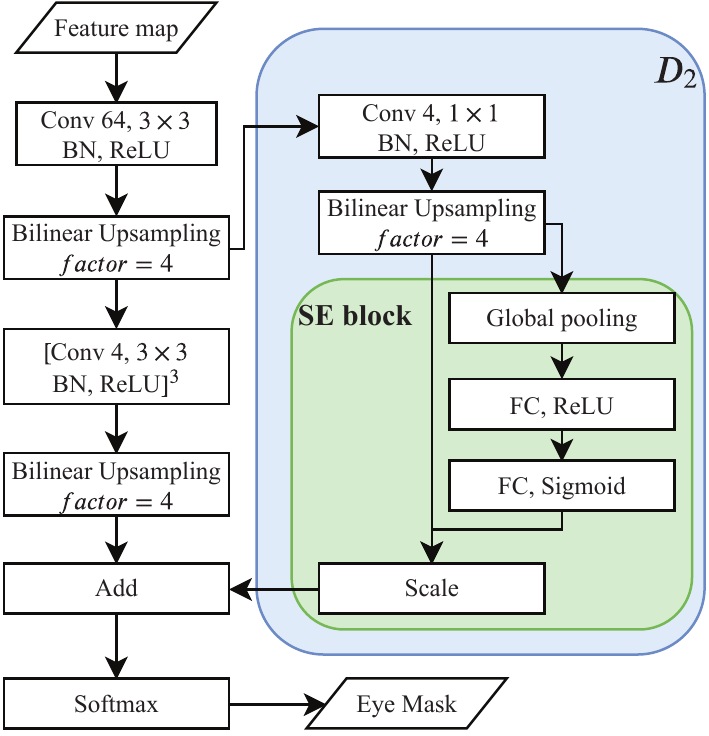}
    \caption{Eye decoder architecture.}
    \label{fig:eye_decoder}
\end{figure}

\subsection{Heuristic filtering}
To reduce the incorrect region in the prediction of the deep network, we analyze the properties of connected components with 8-connectivity. Each mask contains at most four values in $\{0, 1, 2, 3\}$ corresponding to the background, sclera, iris, and pupil. In each class, except the background, we keep only the most massive region which considered as a correct region. Apparently, the sclera covered the iris, and iris wrapped pupil. Consequently, we filled the black holes and removed the small connected components for sclera, iris, and pupil, sequentially as in Algorithm~\ref{alg:filter_ccs}.

\begin{algorithm}[htbp]
    \caption{Filtering with connected components} \label{alg:filter_ccs}
    \hspace*{\algorithmicindent}\textbf{Input}\quad Predicted mask $\mathrm{Mp}$ of size $640\times400$ \\
    \hspace*{\algorithmicindent}\textbf{Output} Filtered mask $\mathrm{Mf}$ of size $640\times400$
    \begin{algorithmic}[1]
        \State Initialize $\mathrm{Mf}$ with zeros
        \For{$i\gets 1, 4$}
            \State $\mathrm{Bw}\gets$ Binary image from $\mathrm{Mp}$ with threshold $i$
            \State $\mathrm{Bw}\gets$ Fill all black holes in $\mathrm{Bw}$
            \State $\mathrm{Lc}\gets$ \parbox[t]{\dimexpr.9\linewidth-\algorithmicindent}{Largest connected component with 8-connectivity in $\mathrm{Bw}$\strut}
            \State $\mathrm{Lr}\gets \mathrm{Bw}\backslash\mathrm{Lc}$
            \State \parbox[t]{\dimexpr\linewidth-\algorithmicindent}{Fill the region in $\mathrm{Mf}$ corresponding with $\mathrm{Lc}$ with the value of $i$\strut}
            
            \State \parbox[t]{\dimexpr\linewidth-\algorithmicindent}{Fill the region in $\mathrm{Mp}$ corresponding with $\mathrm{Lr}$ with $0$ values\strut}
        \EndFor
    \end{algorithmic}
\end{algorithm}

\begin{figure}[htbp]
  \centering
  \begin{minipage}[b]{0.32\linewidth}
    \includegraphics[width=\textwidth]{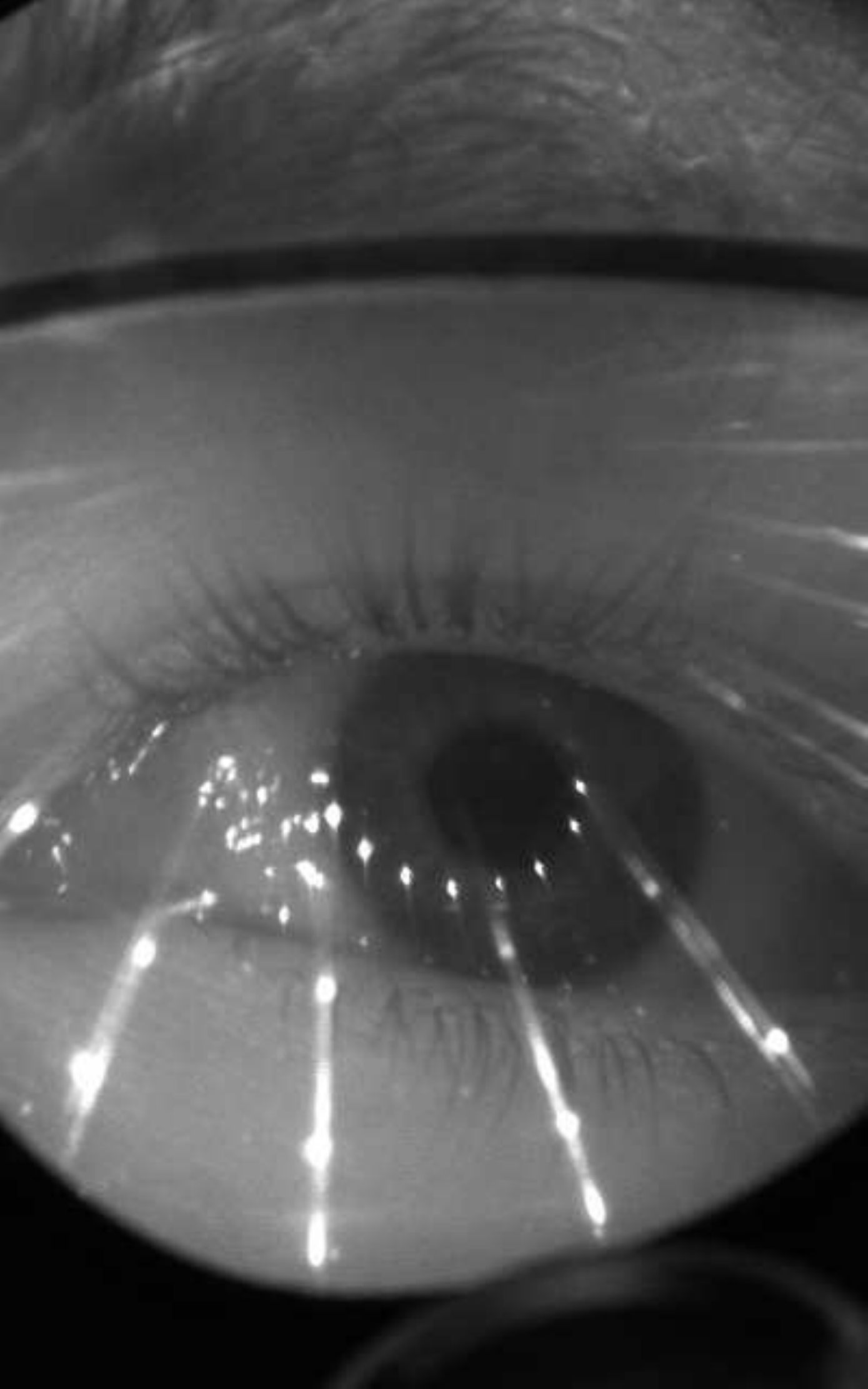}
  \end{minipage}
  \begin{minipage}[b]{0.32\linewidth}
    \includegraphics[width=\textwidth]{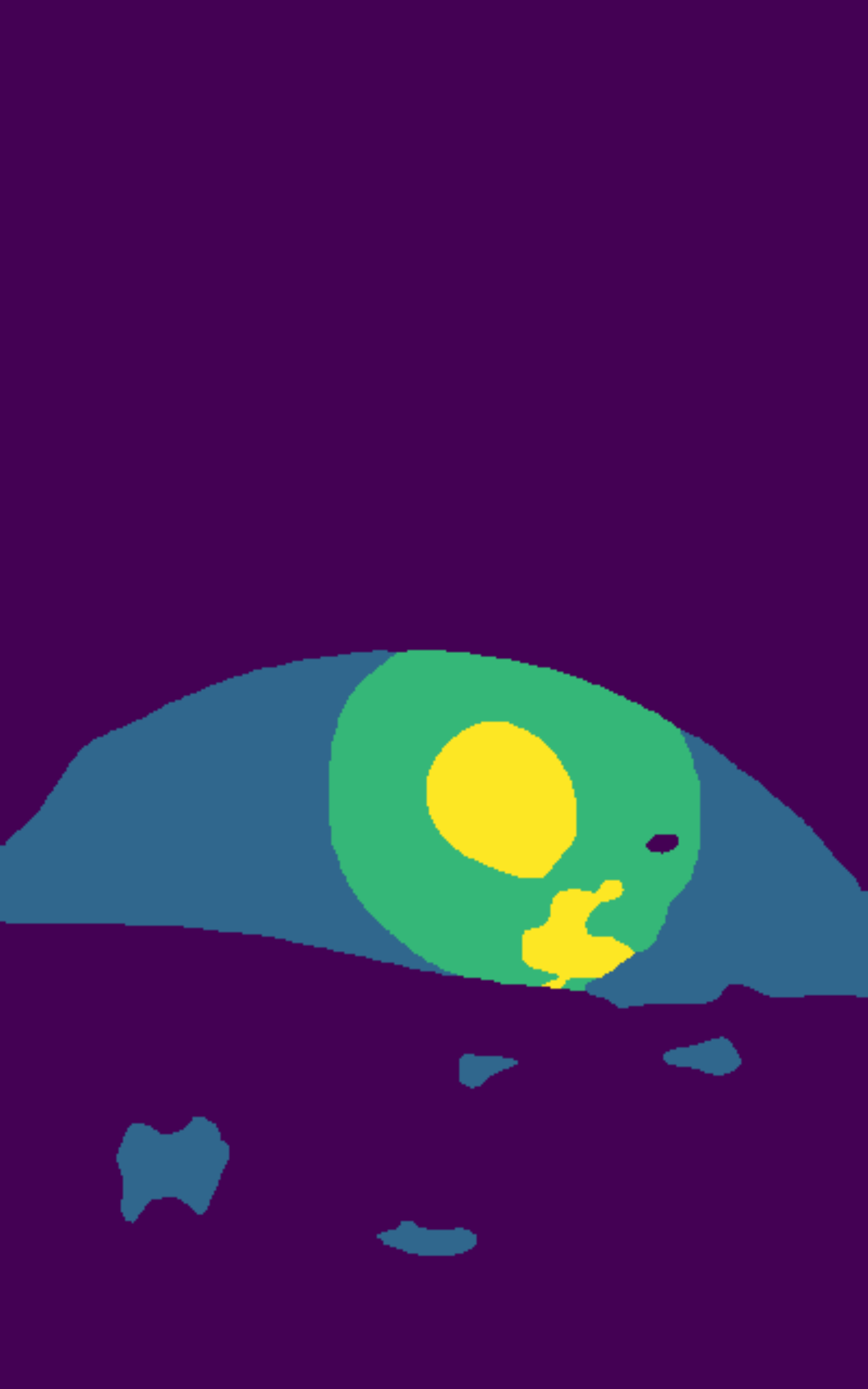}
  \end{minipage}
  \begin{minipage}[b]{0.32\linewidth}
    \includegraphics[width=\textwidth]{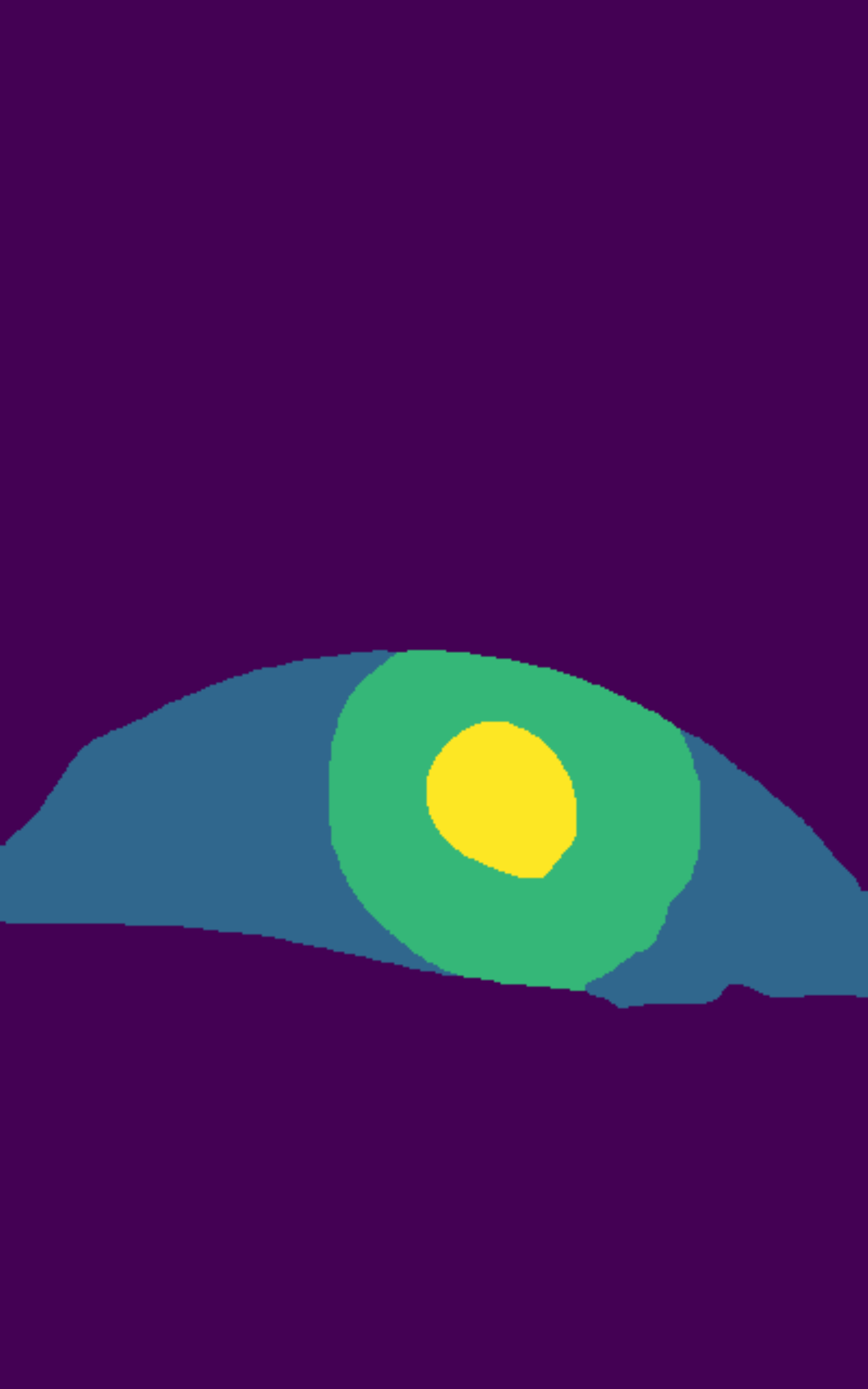}
  \end{minipage}

  \begin{minipage}[b]{0.32\linewidth}
	\includegraphics[width=\textwidth]{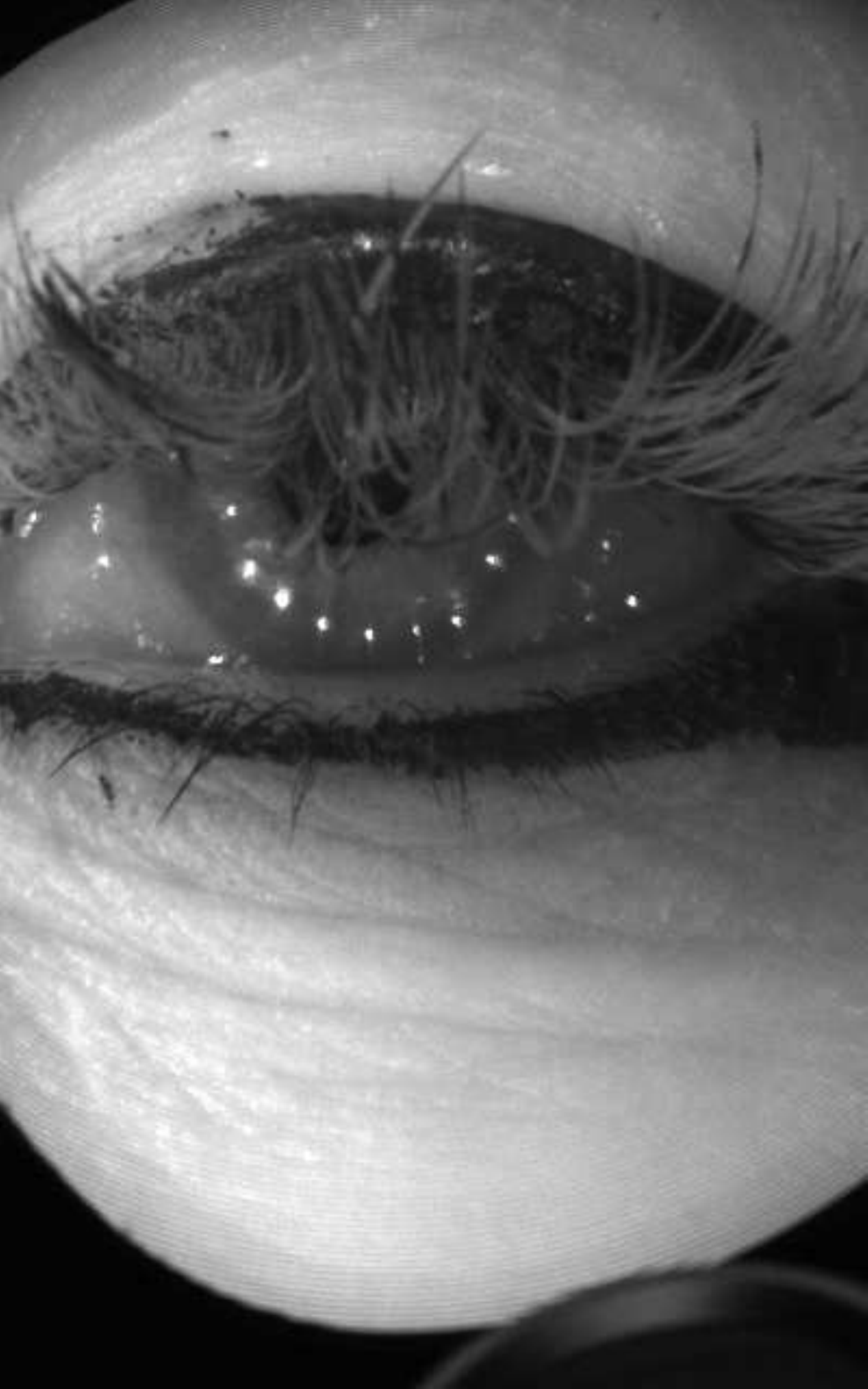}
  \end{minipage}
  \begin{minipage}[b]{0.32\linewidth}
	\includegraphics[width=\textwidth]{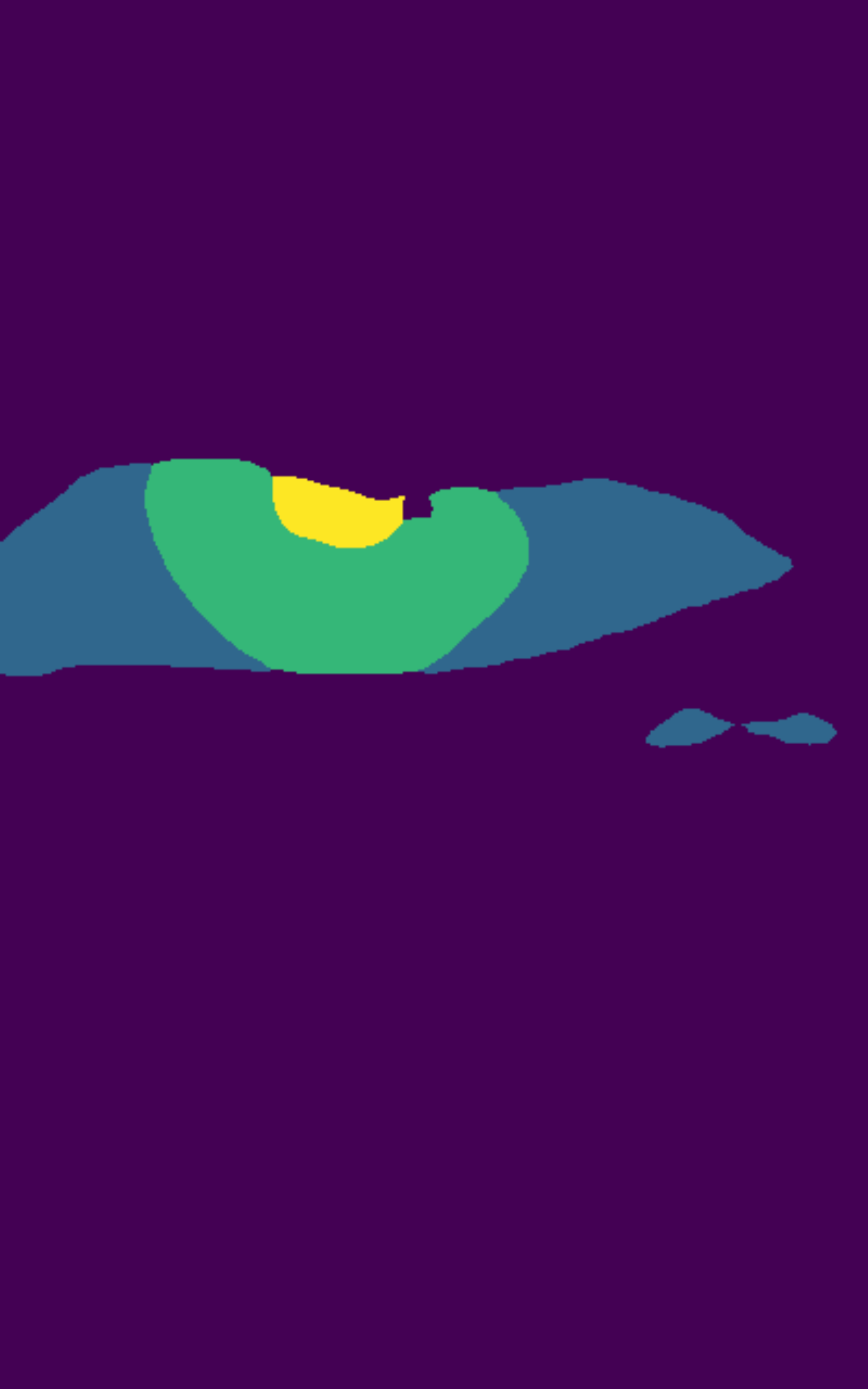}
  \end{minipage}
  \begin{minipage}[b]{0.32\linewidth}
	\includegraphics[width=\textwidth]{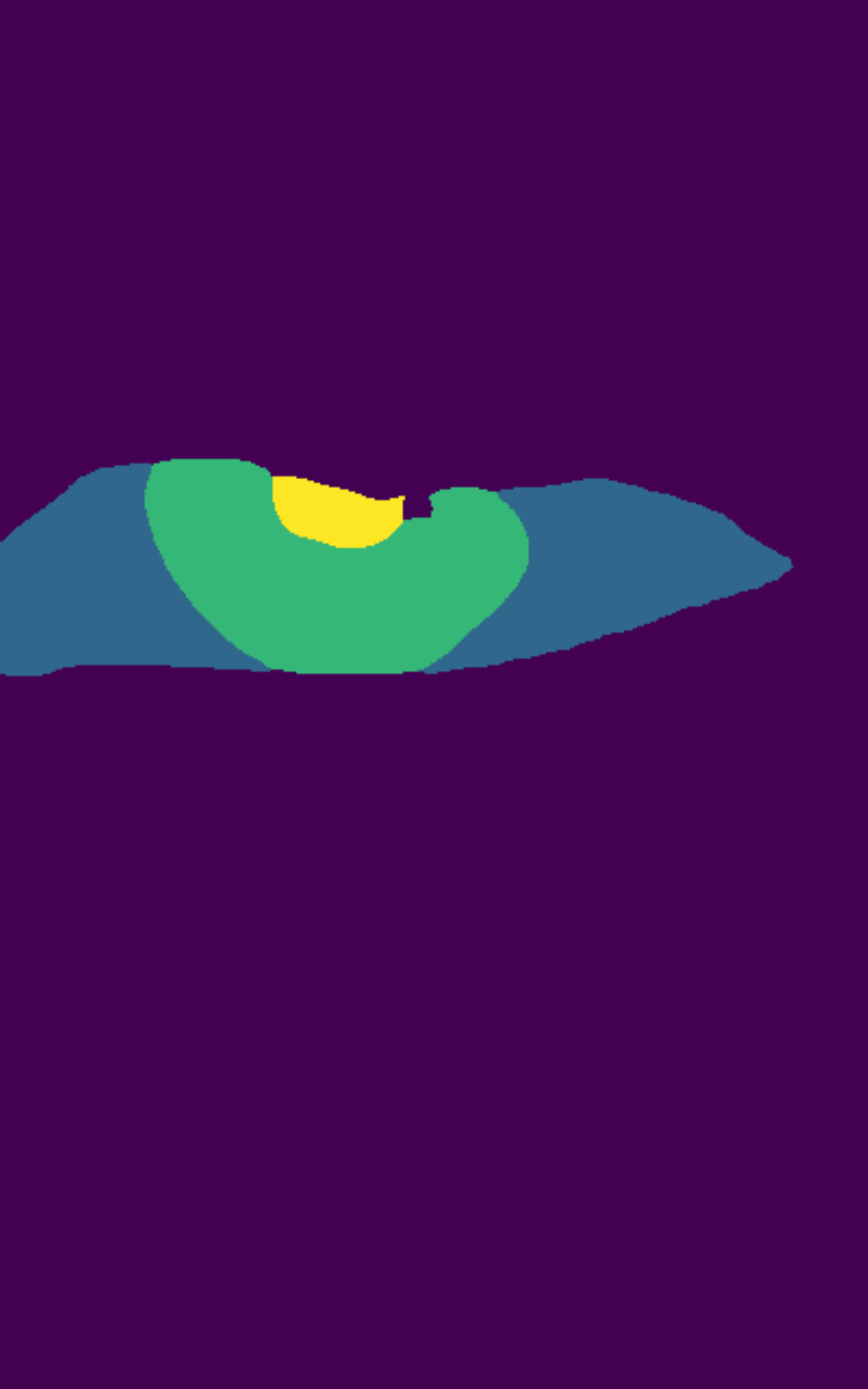}
  \end{minipage}
  \caption{Example of the filtering with connected components. From left to right: input image, predict mask $\mathrm{Mp}$, filtered mask $\mathrm{Mf}$.}\label{fig:heuristic}
\end{figure}

\section{Experimental results}
The experiment and evaluation of our approach were done with Open Eye Dataset (OpenEDS)~\cite{garbin2019openeds} which used in Track-1, semantic segmentation, of OpenEDS challenge. The data were collected from 152 individual participants using a Head-mounted display (HMD) with two synchronized cameras. The semantic segmentation data contained 12759 images which annotated at a resolution of $640\times400$ with three components of an eye: the sclera, iris, and pupil. It divided into three parts for training, validation, and testing with 8916, 2403, 2880 samples, respectively.
\subsection{Implementation}\label{sec:impl}
We implemented our networks in PyTorch~\cite{paszke2017automatic}, and trained for 200 epochs with a combination of Adam optimizer and Stochastic Weight Averaging technique (SWA)~\cite{izmailov2018averaging} from the $51^{th}$ epoch to improve the generalization, weight decay $1\mathrm{e}{-4}$, batch size $32$, initial learning rate $1\mathrm{e}-3$. The weights initialized with He initialization~\cite{he2015delving}. We decreased the learning rate after the $27^{th}$ epoch and kept it as constant value $5\mathrm{e}-4$ during the last 154 epochs. We used the generalized dice loss~\cite{sudre2017generalised} as an objective function for training the network. In the training of each network, we kept only the model which has the least value on the validation loss. We changed the brightness of each image used for training by a factor which was chosen uniformly from $[0.5, 2.0]$.

\subsection{Evaluation}
The following equation~\cite{openedsproposal2019} was used to evaluate the performance of our method
\begin{equation}
    M = 50\left[mIOU + \min\big\{1, \frac{1}{S}\big\}\right]\label{eq:o_metric}
\end{equation}
where
\begin{align}
    \mathrm{mIOU} =& \frac{1}{m}\sum_{i=1}^{m}\frac{|P_{i}\cap G_{i}|}{|P_{i}\cup G_{i}|}\label{eq:p_metric}, \\
    S =& \frac{T \times 4}{1024 \times 1024}\label{eq:s_metric}
\end{align}
with $P_{i}, G_{i}$ are the region of class $i^{th}$ from the ground truth and predicted mask, respectively. The number of trainable parameters of deep network defined by $T$. In other words, the metric $M$ is the combination of the mean intersection over union $\mathrm{mIOU}$ and model size in megabyte $S$.

\begin{figure}[htbp]
    \centering
    \centering
    \begin{minipage}[b]{0.32\linewidth}
        \includegraphics[width=\textwidth]{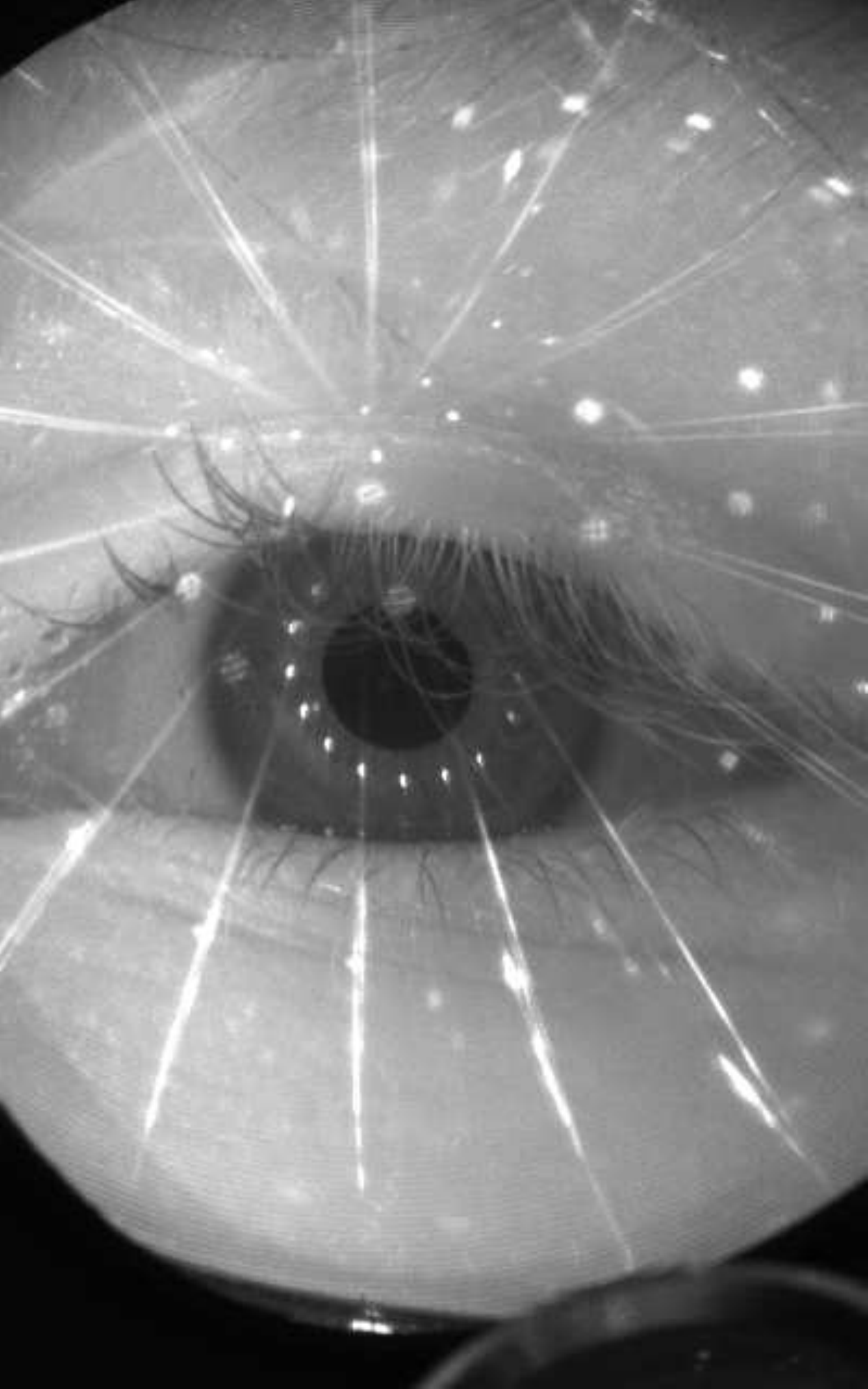}
    \end{minipage}
    \begin{minipage}[b]{0.32\linewidth}
        \includegraphics[width=\textwidth]{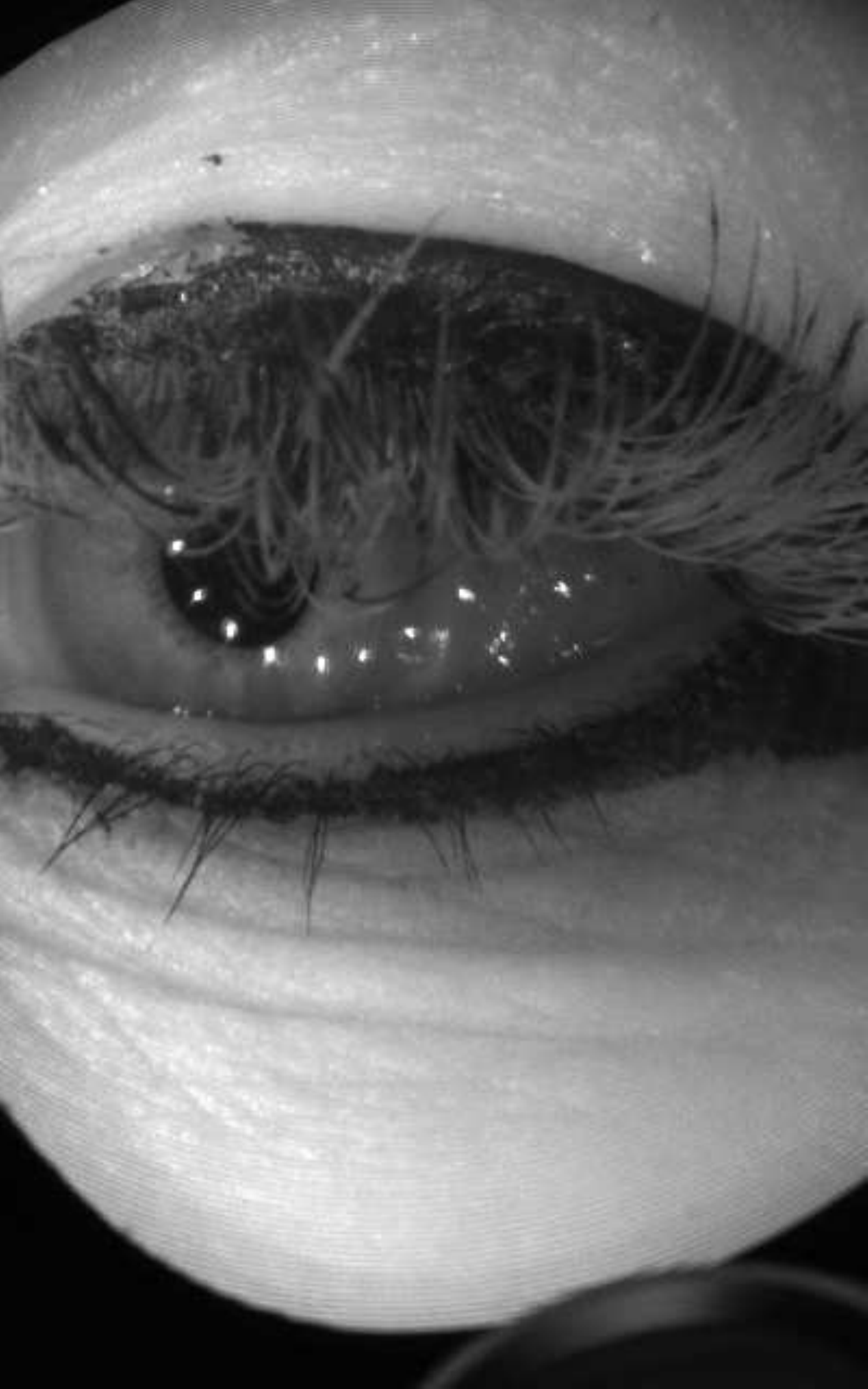}
    \end{minipage}
    \begin{minipage}[b]{0.32\linewidth}
        \includegraphics[width=\textwidth]{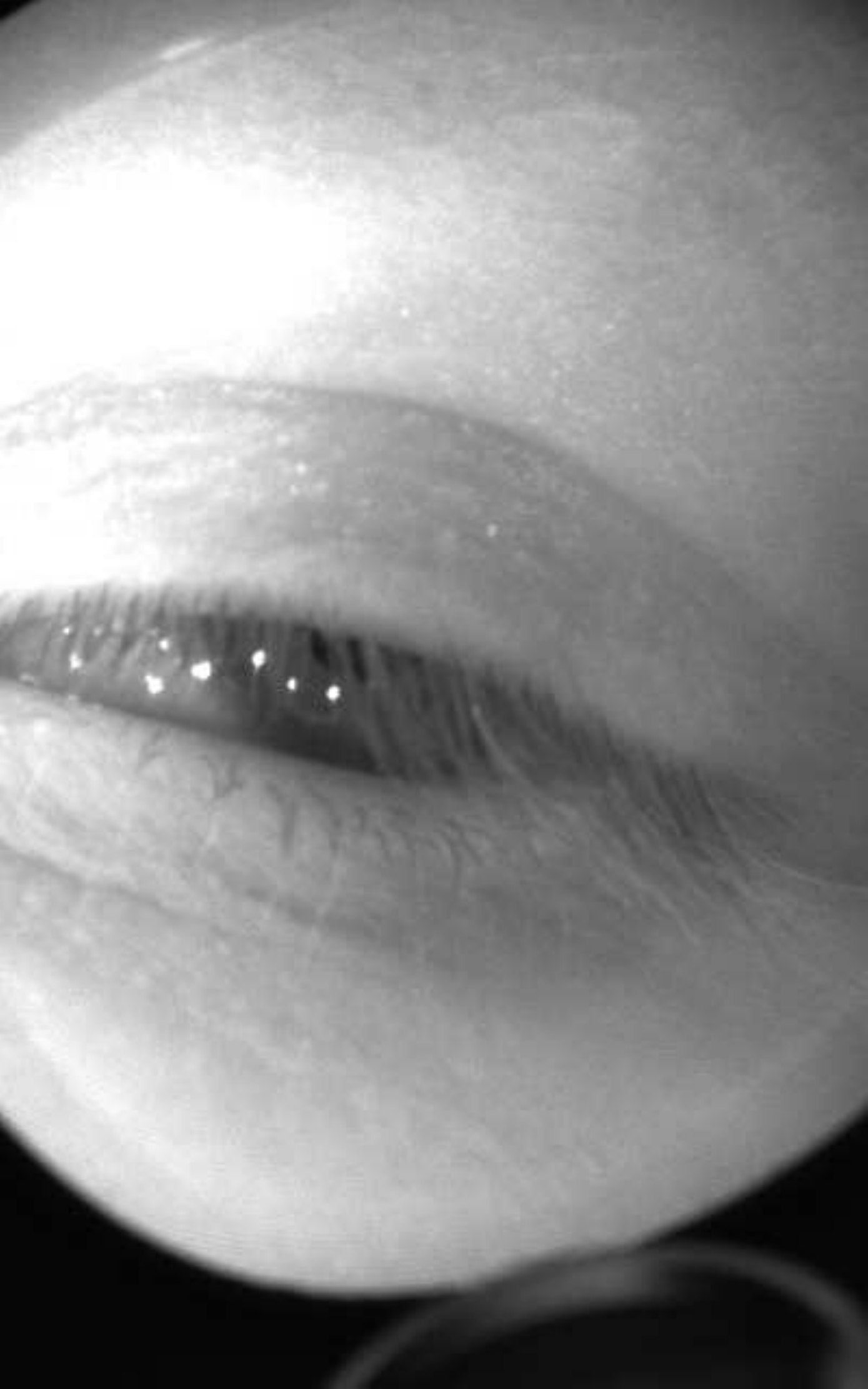}
    \end{minipage}
  
    \begin{minipage}[b]{0.32\linewidth}
        \includegraphics[width=\textwidth]{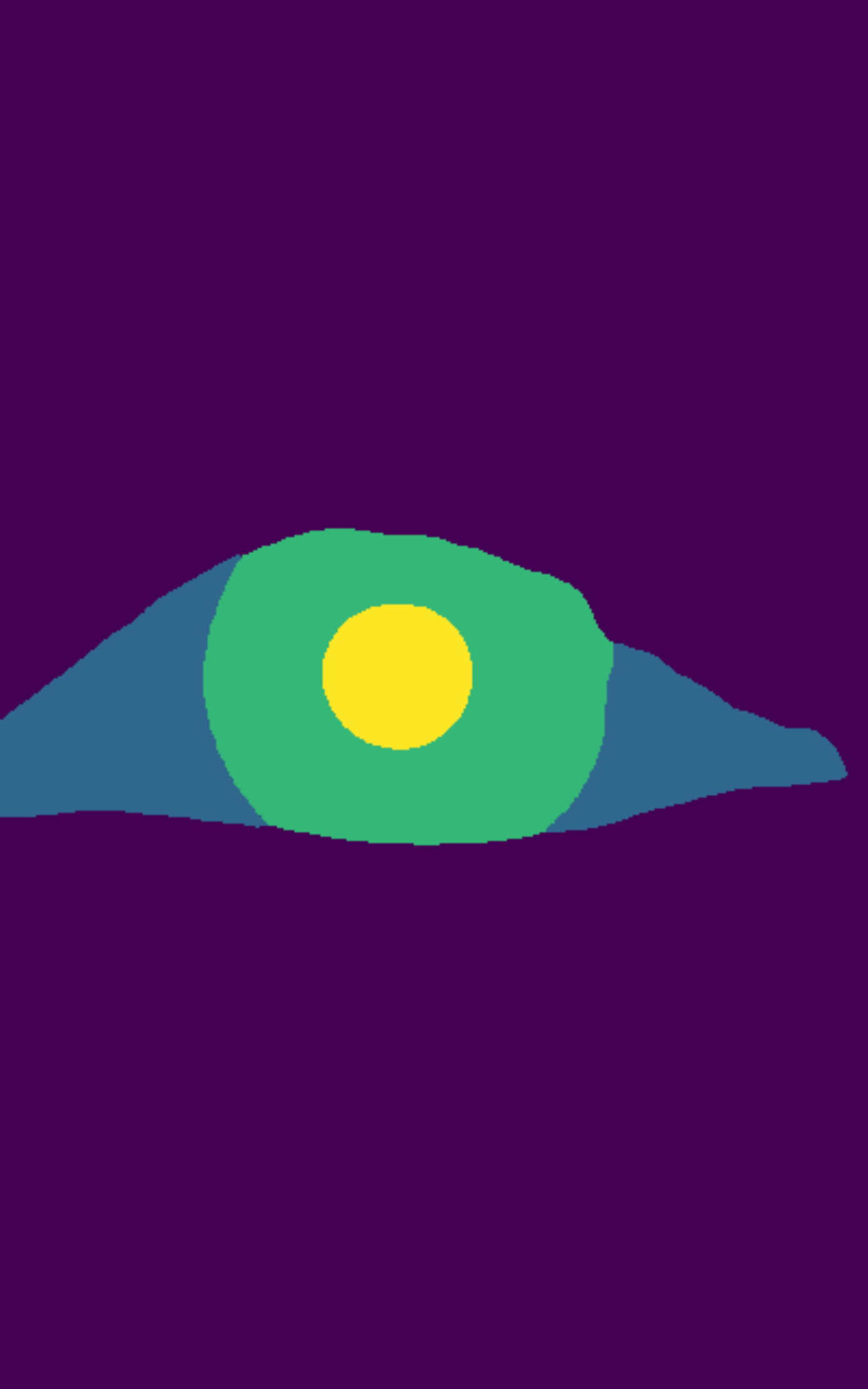}
    \end{minipage}
    \begin{minipage}[b]{0.32\linewidth}
        \includegraphics[width=\textwidth]{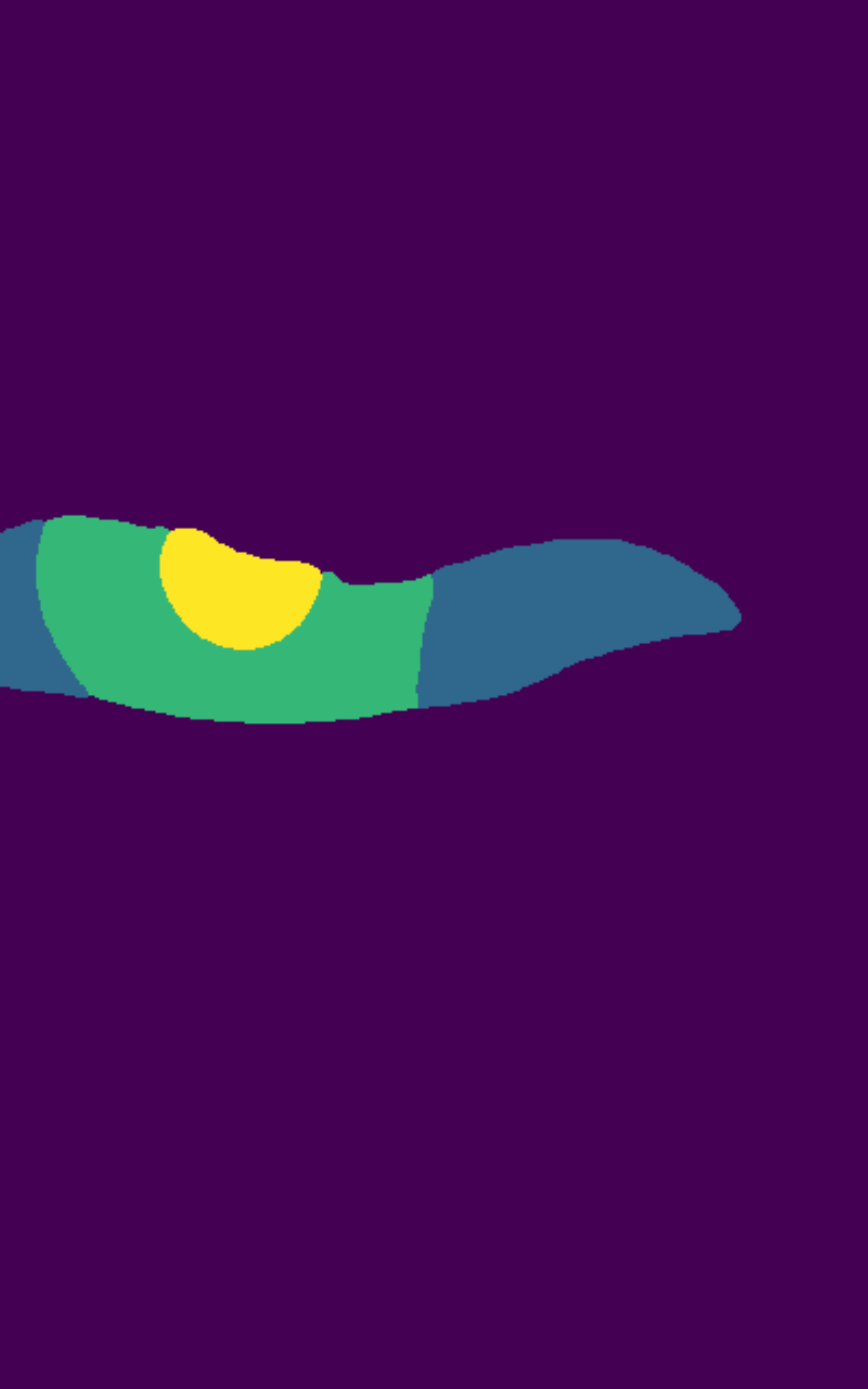}
    \end{minipage}
    \begin{minipage}[b]{0.32\linewidth}
        \includegraphics[width=\textwidth]{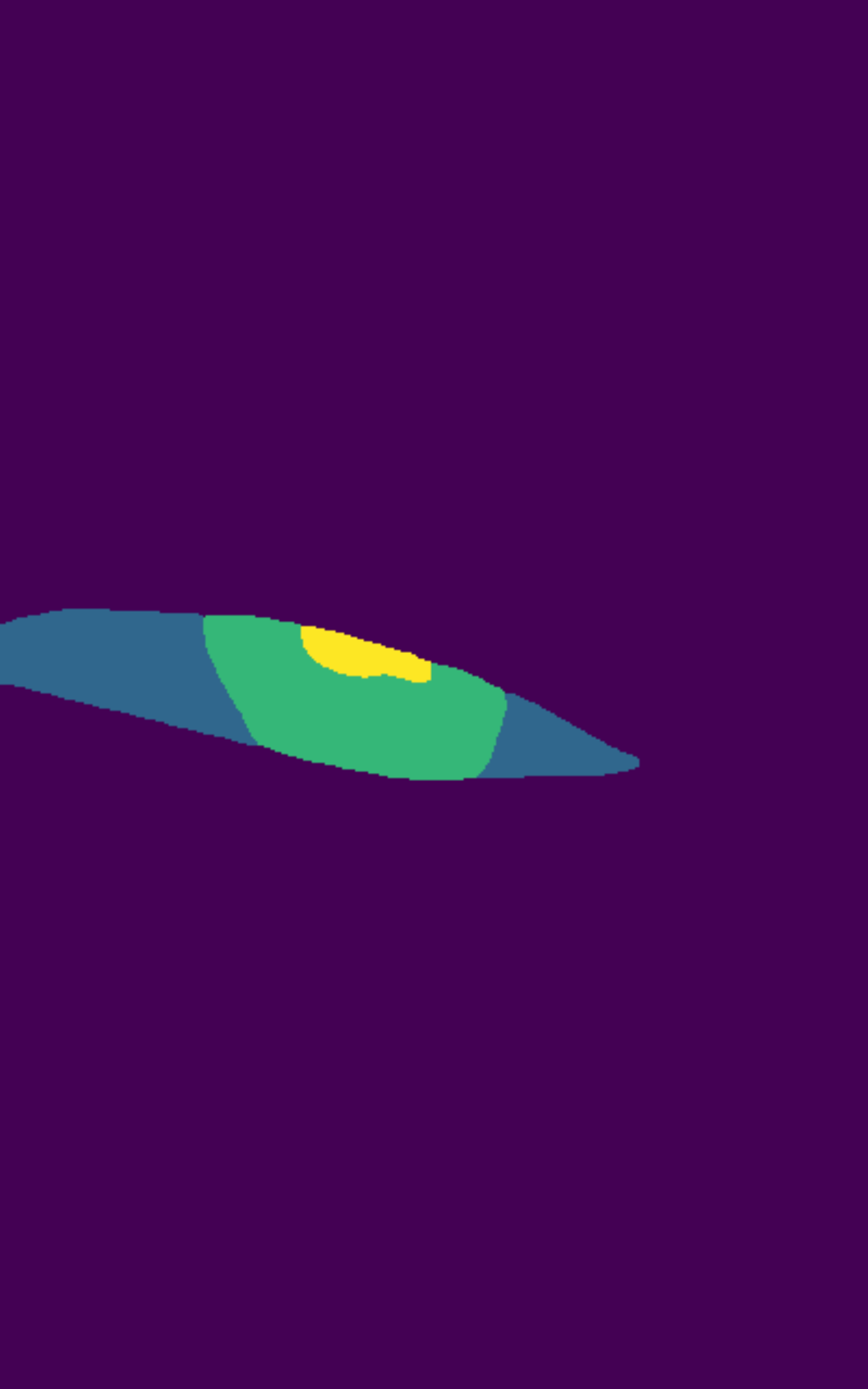}
    \end{minipage}
    
    \caption{Examples on the test set. Input images in the first row and the results of our method in the second row.}
    \label{fig:ex_test}
\end{figure}

\autoref{tab:res_test} shows the results of our method on the test set. The difference between $N_{1}, N_{2}, N_{3}$ is the decoder structure. $N_{3}$ included whole architecture as showed in~\autoref{fig:eye_decoder}. We excluded $D_{2}$, SE block from $N_{3}$ to create $N_{1}$ and $N_{2}$, respectively. All of them are trained from scratch as described in previous. We observed that all models generated the incorrect regions along with the actual eye for eye images with eyeglasses and we fixed that with the heuristic filtering, ~\autoref{fig:heuristic}. Besides that, the heavy mascara which made some eye area invisible also effected our method,~\autoref{fig:ex_test}.

\begin{table}[htbp]
    \begin{center}
    \begin{tabular}{|M{.18\linewidth}|M{.12\linewidth}|M{.14\linewidth}|M{.15\linewidth}|M{.14\linewidth}|} \hline
         Model  & mIOU & \#Params & Model size S & Overall M\hfil \\ \hline \hline
         mSegnet w/ SC~\cite{garbin2019openeds} & $0.895$ & $40000$ & $1.5259$ & $0.7751$ \\
         $N_{1}$ & $0.9482$ & $104456$ & $\mathbf{0.3985}$ & $0.97408$\\
         $N_{2}$ & $0.9483$ & $104720$ & $0.3995$ & $0.97417$ \\
         $N_{3}$ & $\mathbf{0.9485}$ & $104728$ & $0.3995$ & $\mathbf{0.97425}$\\ \hline
    \end{tabular}
	\end{center}
    \caption{Results on the test set. \#Params: the number of learnable parameters. Model size: follow the~\autoref{eq:s_metric}.}
    \label{tab:res_test}
\end{table}

\section{Conclusion}
We presented an effective method for multi-class eye segmentation, which can run on any hardware for real-time inference. In particular, we realized that the area difference between a convex closure of an eye with the boundary curve of the upper part of an eye usually minimal in many cases. It is a crucial component in the future attempts to overcome the missing regions caused by heavy mascara.

\section*{Acknowledgments}
This research was supported by Basic Science Research
Program through the National Research Foundation of
Korea (NRF) funded by the Ministry of Education (NRF-
2017R1A4A 1015559, NRF-2018R1D1A3A03000947).
{\small
\bibliographystyle{ieee_fullname}
\bibliography{./ref/egbib}
}

\end{document}